\documentclass[runningheads]{llncs}
\usepackage{url}
\usepackage{booktabs}
\usepackage{algorithm}
\usepackage{algorithmic}
\usepackage{listings}
\usepackage{amssymb}
\usepackage{amsmath}
\usepackage[utf8]{inputenc}
\usepackage{url}
\usepackage{paralist}
\usepackage{enumerate}
\usepackage{tikz}
\usetikzlibrary{backgrounds}
\usetikzlibrary{calc}
\usepackage{hyperref}

\usepackage{natbib}
\usepackage[english]{babel}

\usepackage{orcidlink}
\usepackage[misc,geometry]{ifsym}

\usepackage{paper}
\usepackage{code}

\usepackage{footnote}
\makesavenoteenv{tabular*}
\makesavenoteenv{table*}
\makesavenoteenv{tabular}
\makesavenoteenv{table}

\usepackage{todonotes}
\setuptodonotes{fancyline}

\usepackage[final]{showlabels}

\begin{document}

\author{Albert Nössig \inst{1,2} (\Letter) \orcidlink{0000-0001-7688-9836} \and
Tobias Hell \inst{2} \orcidlink{0000-0002-2841-3670} \and
Georg Moser \inst{1} \orcidlink{0000-0001-9240-6128}}

\authorrunning{A. Nössig et al.}

\institute{Department of Computer Science, University of Innsbruck,
Tyrol, Austria\\
\email{georg.moser@uibk.ac.at} \and
Data Lab Hell GmbH, Europastraße 2a, 6170 Zirl, Tyrol,
Austria\\
\email{\{albert.noessig, tobias.hell\}@datalabhell.at}\\ 
}
\title{Rule by Rule: Learning with Confidence through Vocabulary Expansion}

\maketitle

\begin{abstract}
In this paper, we present an innovative iterative approach to rule learning specifically designed for (but not limited to) text-based data. Our method focuses on progressively expanding the vocabulary utilized in each iteration resulting in a significant reduction of memory consumption. Moreover, we introduce a \emph{Value of Confidence} as an indicator of the reliability of the generated rules. By leveraging the \emph{Value of Confidence}, our approach ensures that only the most robust and trustworthy rules are retained, thereby improving the overall quality of the rule learning process. We demonstrate the effectiveness of our method through extensive experiments on various textual as well as non-textual datasets including a use case of
significant interest to insurance industries, showcasing its potential for real-world applications.

\end{abstract}

\section{Introduction}
\label{Introduction}

In recent years, the rapid advancement of Artificial Intelligence (AI) technologies has revolutionized various industries and aspects of our daily lives (cf.~\cite{ai_survey, ai_survey2, ai_in_daily_life}, for instance). However, as AI systems become more complex and sophisticated, the need for transparency and interpretability in their decision-making processes has become increasingly crucial. The concept of \emph{Explainable Artificial Intelligence} (XAI; see for example~\cite{xai_review, xai_survey}) has emerged as a response to this demand, aiming to enhance the trust, accountability and understanding of AI systems by providing explanations for their outputs and actions.

Indeed, in many application areas of machine learning, like automotive, medicine, health and insurance industries, etc.,\ the need for security and transparency of the applied methods is not only preferred but increasingly often of utmost importance or even required by law (cf. \emph{EU Artificial Intelligence Act} for instance).

A classical example in this context -- often categorized as \emph{most informative} in the area of XAI~(\cite{hulsen}) --  is the generation of deterministic (if-then-else) rules that can be used for classification. For instance, regarding the prediction of the health status of a patient the easily comprehensible rule shown below is clearly preferable over the unexplainable outcome of a \emph{black-box} like a neural network for both the doctor as well as the patient since the decision is fully transparent.

\begin{lstlisting}[style=Prolog]
		IF	BloodPressure in [70,80]
		  AND	Insulin in [140,170]
		THEN	Diabetes = Yes.

\end{lstlisting}

The field of \emph{Rule Induction}~(\cite{F12}) investigates the construction of simple if-then-else rules from given input/output examples and provides some commonly applied methods to obtain deterministic rules for the solution of a (classification) problem at hand (cf. Section~\ref{RuleLearning}). Representative examples of such rules are shown for each data set considered in our experiments in Section~\ref{Evaluation}, illustrating the major advantages of rule learning methods, namely their transparency and comprehensibility, which make them a desirable classification tool in many areas. 

Unfortunately, these benefits are coupled with the major drawback of generally less accurate results -- often referred to as \emph{interpretability-accuracy trade-off}~(\cite{trade-off}). Moreover, for a long time it has not been possible to efficiently apply rule learning methods on very large data sets~(\cite{mitra18}) as considered for instance in the industrial use case discussed in Section~\ref{UseCase} which is of central interest to us and our collaboration partner -- the \emph{Allianz Private Krankenversicherung (APKV)}. 
We have already extensively investigated these issues in the course of our collaboration with the above-mentioned company from insurance industries with the basic aim to establish rule learning methods -- particularly \foil~(\cite{foil}) and \ripper~(\cite{ripper}) -- as an efficient tool in the reimbursement process. In previous work (\cite{modular_approach, voting_approach}) we introduced approaches to solve the above-mentioned difficulties concerning the application of rule learning methods in a production environment at least to some extent. First, we developed a modular approach (cf. Section~\ref{ModularApproach}) enabling the application of ordinary rule learning methods such as \foil\ and \ripper\ on very large data sets including several hundreds of thousands examples. 
However, the in general poorer performance compared to state-of-the-art methods with respect to accuracy remained. So, we came up with an extension of the introduced modular approach in the form of the voting approach shortly described in Section~\ref{VotingApproach}. After consultation with our collaboration partner, we agreed that at the end of the day it is even more important to ease the understanding of a classification made than to make the whole procedure fully transparent. So, this additional step in the process of decision making deals with the \emph{interpretability-accuracy trade-off} by incorporating an ensemble of explainable as well as unexplainable methods. As a consequence, the procedure loses its full transparency but gains a significant improvement of classification accuracy, while preserving \emph{end-to-end explainability} by corroborating each prediction with a comprehensible rule.

At this point we have already made a huge step towards the application of trustworthy AI methods in the company. However, another crucial problem that is not solved in a satisfying manner by the combination of the two approaches above is the handling of text-based data. The data basis for the reimbursement use case is a collection of (scanned) bills where we extracted the most important information in the form of nominal (and continuous) attributes as described in more detail in Section~\ref{UseCase}. Unfortunately, by this way of preprocessing we might lose a lot of additional information given by the original textual data.

However, up to this point, we have mainly considered nominal data with the only exception of the \emph{IMDB movie reviews} data set\footnote{See \url{https://www.kaggle.com/datasets/lakshmi25npathi/imdb-dataset-of-50k-movie-reviews}.} which has been part of the benchmark data sets in the evaluation of our modular approach. The results have not been really satisfying because the achieved accuracy has been below our expectations on the one hand -- which is solvable by our voting approach at least to some extent -- but on the other hand it has shown that the form and complexity of the generated rules is not reasonably applicable for (end-to-end) explainable classification. What seems to be not too problematic considering the comparatively small IMDB data set, is the choice and especially the size of the underlying dictionary used to generate rules. For the movie reviews we simply considered the thousand most common words in the data set but the bills handed in to the insurance are far more complex. They usually consist of at least one page of text using partly highly complicated technical terms from various medical fields instead of 2-3 sentences describing personal opinions about movies in simple language. Note that simply using a much larger dictionary as basis for the rule learning process is not the remedy because the computation time as well as the memory consumption for the generation of the rules increases drastically with increasing dictionary size. In this paper, we especially aim to gain more control over the complexity of the generated rules and make it possible to reasonably apply rule learning methods such as \foil\ and \ripper\ also on text-based data by starting off with a concise dictionary (designed by domain experts) and decreasing the number of considered examples before extending the applied dictionary in an iterative way. The intention behind this approach is to learn \emph{general} rules in a first step using a small and computationally rather cheap dictionary for a very large number of input examples. With each learned rule the number of considered positive examples decreases by definition of the rule learning algorithms. When a certain point is reached -- either a predefined number of iterations or a condition regarding the quality of a rule as described in detail in Section~\ref{Methodology} -- we extend the dictionary to handle more \emph{specific} examples. This way of proceeding can be repeated until a quite comprehensive dictionary is applied on a few remaining \emph{edge cases}.    
In addition, the basic idea behind this way of proceeding can be applied also on nominal (and continuous) data in order to improve the \emph{quality} of a rule as explained in Section~\ref{Methodology} and shown in the experimental evaluation in Section~\ref{Evaluation}.

Apart from evaluating our approach on common benchmark data sets regarding classification of textual data (\emph{IMDB}~(\cite{imdb}), \emph{Reuters-21578}~(\cite{reuters-21578}), \emph{Hatespeech}\footnote{See \url{https://www.kaggle.com/datasets/mrmorj/hate-speech-and-offensive-language-dataset}.}), we also show the advantages of the basic idea of our approach applied on non-textual data considering some common data sets from the \emph{UCI Machine Learning Repository}~(\cite{Dua:2019}) or \emph{kaggle}\footnote{See \url{https://www.kaggle.com/}.}, respectively. Moreover, we present novel results on explainable \emph{classifications of bills for reimbursement} particularly using textual data as input. The latter case study stems from an industrial collaboration with \emph{Allianz Private Krankenversicherung (APKV)} which is an insurance company offering health insurance services in Germany.

Summed up, our main goal is to solve a text-based classification problem in reasonable time and computational complexity by applying easily comprehensible rules that have been generated by using a dictionary of variable size.  
Moreover, we define a measure for the quality of a rule and integrate it in the iterative way of proceeding our proposed approach is based on. As shown in the experiments, this iterative rule refinement is beneficial even for non-textual data.
All in all, this paper directly builds on our previous work and expands upon the approaches presented therein to handle especially textual data more efficiently and gain more control over the complexity of the generated rules by iteratively extending the size of the applied dictionary (or in general the number of attributes).

More precisely, we make the following contributions.
\subsubsection{Iterative Approach Based on Rule Learning} We introduce a novel iterative approach based on rule learning exploiting the benefits of a variable number of attributes (in particular an adaptable dictionary) during the generation of a rule set (see Section~\ref{Methodology} for further details). 

Together with the modular as well as the voting approach introduced in our previous work (\cite{modular_approach, voting_approach}), this makes rule learners a serious alternative to state-of-the-art classification tools and enables the application of tried and trusted rule learning methods in a complex and diverse production environment.

\subsubsection{Experimental Evaluation} Further, we provide ample experimental evidence that our methodology not only clearly simplifies the application of rule learning methods on text-based data but also provides significant improvements on the accuracy for the standard benchmarks (see Section~\ref{Evaluation}).

\subsubsection{Industrial Use Case} Finally, we show that our approach makes it possible to efficiently apply the way of proceeding we successfully introduced in previous work now also on text-based data, in particular the raw OCR scans used for reimbursement.
We emphasise that our classification yields comprehensible rules that are of direct interest to our industrial collaboration partner (see Section~\ref{UseCase}).

\paragraph*{Overview.}    

In Section~\ref{Notations} we introduce the major definitions and notations as well as the general ideas behind our approaches from previous work. 
Section~\ref{RelatedWork} serves to discuss related work focusing on similar goals as considered in this paper, especially on various forms of (explainable) text-based classification, while we concretely introduce our aforementioned iterative approach as well as the \emph{Value of Confidence} applied therein as a measure of reliability of a rule in Section~\ref{Methodology}. Section~\ref{Evaluation} provides ample evidence of the advantages of our approach
and presents the case study mentioned. Finally, in Section~\ref{Conclusion} we summarize the main results and discuss ideas for future work.

\section{Notations \& Preliminaries}
\label{Notations}

After motivating the basic idea behind the approach introduced in this paper, we give a more comprehensive summary of rule learning in general as well as the work we have already done in this field in this section.

\subsection{Rule Learning}
\label{RuleLearning}

As already mentioned in the introduction, the field of \emph{Rule Induction} focuses especially on providing efficient algorithms for the generation of simple if-then-else rules as we are mainly interested in. \emph{Repeated Incremental Pruning to Produce Error Reduction} (RIPPER; \cite{ripper}) is state-of-the-art in this field and, consequently, we consider mainly this algorithm in our experiments. 

However, there are also other fields like \emph{Inductive Logic Programming} (ILP; cf.~\cite{ilp_survey}) that encompass methods yielding results that can be interpreted as if-then-else rules. Basically, ILP-tools investigate the  construction of first- or higher-order logic programs.
In the context of this paper, it suffices to conceive the learnt hypothesis as first-order Prolog clauses as depicted below.
\begin{lstlisting}[style=Prolog]
  H :- L1, ..., Lm
\end{lstlisting}
Here, the \emph{head} $H$ is an atom and the \emph{body} $L_1, \dots, L_m$ consists of literals, that is atoms or negated atoms.  

Consequently, ILP is often conceived as a subfield of inductive programming. However, our
interest stems from the fact that logic programs are (by definition) nothing else but sets of clauses, that is, rules. 

Concerning ILP, especially one of the first tools from this field, the \foil\ algorithm (\emph{First Order Inductive Learner}; \cite{foil}), is of main interest to us due to its simplicity. In previous work (\cite{modular_approach}), we have extensively investigated also some more modern ILP-tools and in the course of this we have shown that they are mostly not suited for our needs since they are rather designed to generalize from a very small set of input examples. Nevertheless, our \emph{Python} reimplementation of the \foil\ algorithm presented in our previous work is able to handle large data sets straight away which is in particular beneficial for the data set considered in our case study. Moreover, contrarily to more modern ILP-tools which rather aim to generate complex (recursive) programs, \foil\ is very well suited to learning simple if-then-else rules as we want to generate.

Moreover, also the outcomes produced by decision trees (cf. for instance~\cite{trees}) can obviously be interpreted as if-then-else rules. However, this work focuses on the approaches mentioned above since the way of proceeding of trees is not really suited for the ideas introduced in this paper.\\

So, all in all, there is a large variety of methods that can be applied within the iterative approach introduced in this paper. However, in the following we mainly focus on \foil\ and \ripper\ since these two methods have been especially investigated in our previous work as outlined in the following.

\subsection{Modular Approach}
\label{ModularApproach}

\tikzstyle{txtlabel} = []
\tikzstyle{method} = [shape=rectangle, rounded corners, draw, fill=white, minimum width=2.2cm, minimum height=1.2cm]
\begin{figure}[b!]
\centering
\begin{minipage}{\linewidth}
\begin{center}
  \begin{tikzpicture}[scale=.8]
  \begin{scope}
    \node[method] (represent) at (0,0) {
      \begin{minipage}{2.5cm}
        \textbf{\small Representation Learning}
      \end{minipage}};
    
    \ibox{(0,-1.5)} {2.2}{1}{\strut{}...};
    \ibox{(0.1,-1.9)}  {2.2}{1}{\tiny Neural Networks};
    \ibox{(0.2,-2.3)}  {2.2}{1}{\tiny UMAP};
  \end{scope}
  \begin{scope}
  \node[method] (clustering) at (4cm,0) {
    \begin{minipage}{2.5cm}
      \textbf{\small Input Selection}
    \end{minipage}
  };
    \ibox{(4,-1.5)} {2.2}{1}{\strut{}...};
    \ibox{(4.1,-1.9)}  {2.2}{1}{\tiny $k$-means};
    \ibox{(4.2,-2.3)}  {2.2}{1}{\tiny DBSCAN};  
  \end{scope}
  \begin{scope}
  \node[method] (learner) at (8cm,0) {
    \begin{minipage}{2.5cm}
      \textbf{\small Rule Learner}
    \end{minipage}
  };
   \ibox{(8,-1.5)} {2.2}{1}{\strut{}...};
    \ibox{(8.1,-1.9)}  {2.2}{1}{\tiny \foil};
    \ibox{(8.2,-2.3)}  {2.2}{1}{\tiny \ripper};
  \end{scope}
  \node[minimum width=1.3cm,anchor=west,align=left] (answer) at ($(learner.east)+(4mm,0)$) {    
\begin{lstlisting}[style=prolog,basicstyle=\color{darkviolet}\scriptsize\ttfamily,frame=none]
target(V) :-
  black_1(V),
  black_N(V).
\end{lstlisting}
  };
  \begin{scope}[on background layer]
    \draw[->] (represent) -- (clustering);
    \draw[->] (clustering) -- (learner);
    \draw[->] (learner) -- (answer);
  \end{scope}
\end{tikzpicture}
\end{center}
\end{minipage}
\vspace{2ex}
\caption{\textbf{Modular Approach to Rule Learning.} The first phase (\emph{Representation Learning}) is intended to yield a compact representation of the original (high-dimensional) input data. This is advantageous for clustering applied subsequently during the second phase (\emph{Input Selection}). These two steps put in front of the application of a chosen \emph{Rule Learner} in the final phase make it possible to find comprehensible rules on very large data sets in reasonable time.}
\label{fig:mod_approach}
\end{figure}
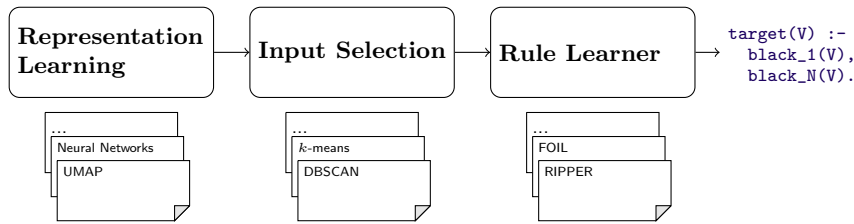

The first problem we have faced regarding the application of rule learning methods in our reimbursement use case has been the (nearly) infeasible complexity caused by the vast amount of examples contained in the corresponding data set. As extensively discussed in the corresponding paper (\cite{modular_approach}), both the time as well as the memory consumption increase drastically with increasing number and length (i.e., number of attributes) of input examples. In order to solve this problem we introduced a \emph{modular approach} that is basically composed of three independent phases as depicted in  Figure~\ref{fig:mod_approach}. The core idea is to make the approach as versatile as possible by allowing to apply a huge variety of methods within each step depending on the kind of input data considered. 

First, an appropriate feature extraction or dimensionality reduction method such as a neural network, \emph{UMAP} (\cite{umap}) or a principal component analysis is applied with the goal to find a compact representation of the high-dimensional input data. This representation should be beneficial for clustering applied in the second step, where a chosen method like \emph{k-means} or \emph{DBSCAN} (\cite{dbscan}), for instance, divides the whole set of input data into various subsets of similar examples. This crucial step of our modular approach is applied on the positive and negative examples separately because at the end of the day we aim to identify very similar positive examples as well as a concise subset of inhomogeneous negative examples representing the whole negative examples contained in the input data set. The idea behind this step is to reduce the complexity of the problem. On the one hand we significantly reduce the number of negative examples to a subset of as heterogeneous examples as possible and on the other hand we exploit the reduced complexity of the feature space resulting from the clustering of similar positive examples as explained in detail in our previous work (\cite{modular_approach}).

The set of negative representatives is concatenated to each cluster of similar positive examples resulting in several independent sets of examples each serving as input for a rule learner such as \foil\ or \ripper\, for example, applied subsequently in parallel in the third and final step. However, note that the rule learner uses the data in its original form instead of the features learned in the first step because otherwise explainability would be lost by generating rules considering incomprehensible features. The rules generated on each subset of input examples are afterwards concatenated to one rule set with the label of the positive examples as target.

Summed up, this approach makes it possible to apply classical rule learning methods on very large data sets in reasonable time without negatively affecting the resulting accuracy. However, the classification accuracy achieved in our experiments was still not satisfying directly confronting us with the next problem, the \emph{interpretability-accuracy trade-off}.

\subsection{Voting Approach}
\label{VotingApproach}

\begin{figure}[h]
\begin{center}
  \setlength{\fboxsep}{0pt}
  \fbox{
    \includegraphics[width = 0.75\textwidth]{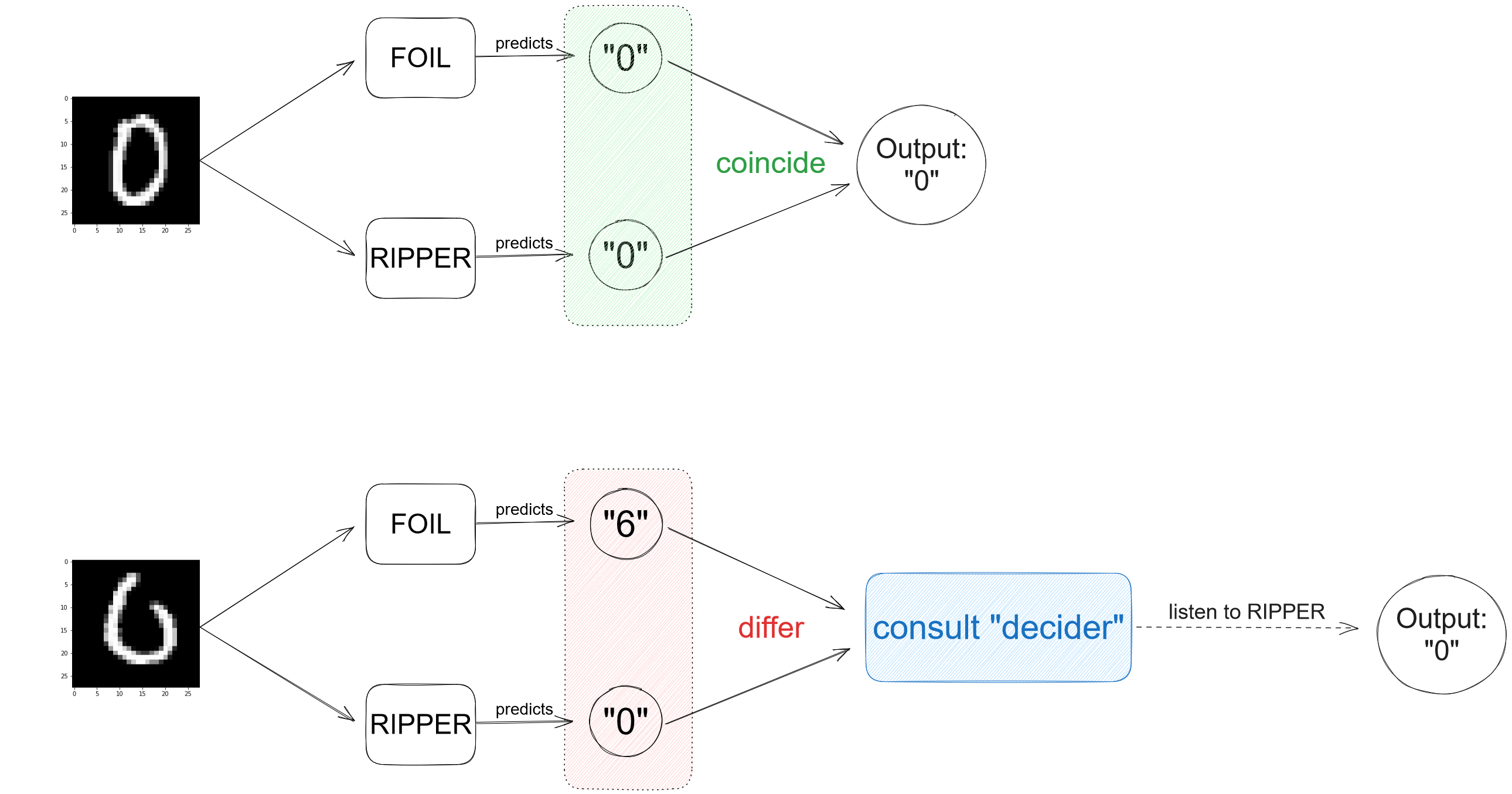}
  }
\end{center}
\caption{\textbf{Voting Approach for end-to-end Explainable Classification.} Generally the approach distinguishes between two basic scenarios, namely coinciding predictions given by the rule learners as well as conflicting ones. In a first step only the explainable methods are considered using the corresponding prediction in case they match. Otherwise, an (unexplainable) state-of-the-art method -- the so-called \emph{decider} -- is consulted to resolve the existing rule conflict.}
\label{fig:voting_approach}  
\end{figure}

As a remedy for the issue of generally less accurate results achieved by explainable methods, we decided to apply an ensemble of classification models consisting of explainable as well as unexplainable methods in a novel kind of voting approach depicted in Figure~\ref{fig:voting_approach} and explained in detail in the corresponding paper (\cite{voting_approach}).

The principal idea is to directly build upon the modular approach outlined in Section~\ref{ModularApproach} and make use of the generated rule sets produced therein. We use especially \foil\ and \ripper\ as representative examples since these two algorithms have been mainly used in the predecessor paper but basically they can be replaced by any method yielding if-then-else rules (or something similar that can be transformed into such rules).

In the first step our ensemble of classification models only contains the two explainable methods and we check whether the applied models predict the same class or not. In case the predictions coincide, we directly output the corresponding label corroborated by one rule from each method. In case of different predictions, we additionally incorporate the state-of-the-art prediction from an unexplainable method. Simply put, this method -- the so-called \emph{decider} -- tells us which rule learner is right and we use the according prediction as final classification again confirmed by the rule from the corresponding explainable method. Otherwise, if none of the rule learners predicts the same class as the decider, we do not give a prediction but say that there is no convincing justification for the prediction.

At the end of the day, we have to live with the trade-off that the full transparency of the classical methods is replaced by \emph{end-to-end explainability} meaning that the final classification is justified by an easily comprehensible rule while the steps in between can be supported by the superior performance of an unexplainable (decider) method. However, the trust as well as a basic understanding of the underlying model is still assured and this way of proceeding yields a significant boost of classification accuracy as shown in detail in the paper (\cite{voting_approach}).

So, all in all, in our previous work we have made it possible to apply classical rule learning methods in reasonable time on very large data sets with a significantly improved accuracy compared to the base method. However, up to this point, we have only considered data sets consisting of nominal data. Despite the improvements achieved with the combination of the two approaches introduced above, the application of classical rule learning methods on text-based data sets is still not straightforward, especially concerning the choice of the applied dictionary or feature set, respectively. In some first experiments considering the IMDB data set, we have used the thousand most common words in the data set as features. While the computational complexity of this choice of dictionary is manageable, the corresponding results are not satisfying. On the other hand, applying all words occurring in the data set as features, the computational complexity becomes infeasible. Instead of searching for a dictionary yielding a trade-off between computational costs and classification accuracy, we aim to iteratively adapt the size of the dictionary. In the first iterations, we learn simple rules on a concise dictionary as long as the positive and negative examples are highly different from another such that they can be distinguished by some crucial key words. As soon as the difference gets more subtle (measured by the defined \emph{Value of Confidence} (cf. Definition~\ref{voc}) of the generated rule), we extend the dictionary. For instance, we could double the size of the dictionary and consider the two thousand most common words in the data set. 

This iterative adaptation of the applied dictionary can basically be incorporated in step 3 (\emph{Rule Learning}) of our modular approach additionally increasing the application area of classical rule learning methods. More details are given in Section~\ref{Methodology}.

\section{Related Work}
\label{RelatedWork}

After motivating the basic idea behind the approach introduced in this paper and introducing the concepts applied therein, in this Section we discuss related work that also focuses especially on the (explainable) classification of textual data as well as novel ideas in the context of rule learning in general.\\

Regarding text classification in general there is a huge number of methods out there dealing with this problem. Some surveys summarizing the most common (explainable as well as unexplainable) approaches have been done in recent years for instance by \cite{text_classification_survey1, text_classification_survey2, text_classification_survey3, text_classification_survey4}. Moreover, \cite{rationalisation_survey} have recently published a survey comparing different rationalisation approaches in the context of explainable text classification. 
Furthermore, \cite{semantic_classification} give an overview of common semantic text classification methods and discuss the benefits of these methods over traditional text classification approaches.\\

A more specific method utilizing similar ideas as we apply in our approach is proposed by \cite{KitCat} who introduce a \emph{tool kit for text categorization} called \emph{KitCat} and not only focus on the explainable classification of textual data but also make use of a confidence measure for dealing with ambiguities similar to our \emph{Value of Confidence} introduced in Section~\ref{Methodology}. For evaluation, they consider in particular the \emph{Reuters-21578} data set where they report a micro-averaged precision/recall of $83.8$\%. 
As opposed to their idea of deriving symbolic rules from decision trees that have been optimized to handle in particular sparse data, we directly obtain rules from classical rule learning methods focusing especially on the complexity of the generated rules with respect to the underlying dictionary in order to improve the versatility of the classical methods. 
Note that we cannot really compare the achieved results, since we used a different data split. However, on \emph{NLTK's Reuters corpus}  we report an accuracy of about $80.5$\% and $81.7$\% on \ripper\ and \foil, respectively.

The \emph{Reuters-21578} data set is a common benchmark for the evaluation of various classification methods on text-based input data and has been intensively investigated for instance by \cite{reuters_difficulties}. Another approach from the field of explainable artificial intelligence that considers this data set among others is \emph{Olex-GA} by \cite{olex-ga}. The results of this genetic algorithm are very similar to the \emph{if-then-else} rules generated by the rule learning methods considered by us. In the course of their evaluations, they compare their method among others also with \ripper\ and report comparative but slightly worse classification results considering the \emph{break-even point} -- the average of precision and recall where the difference between them is minimal -- as accuracy metric.

In addition, we consider the \emph{IMDB movie reviews} data set in our experiments which has been investigated also by \cite{imdb_nlp_approach}, for instance, who utilize ideas from neuro-symbolic learning (cf.~\cite{neuro_symbolic_ai}) in a semi-supervised machine learning approach resulting in interpretable results in the form of linear combinations of attention scores. They report remarkable results of a F1-score of up to $89.41$\% but apparently they used a subset or a different version of the data set we used in our experiments since they consider a total amount of 25 thousand examples compared to the 50 thousand examples used by us resulting in a F1-score of $76.5$\%.
Moreover, regarding this approach it should be noted that there is an ongoing discussion concerning the interpretability of attention weights (cf.~\cite{attention_explanation1, attention_explanation2}), whereas the \emph{if-then-else} rules generated by the rule induction methods applied in our approach are commonly categorized as \emph{most informative} in the area of XAI.\\

Regarding the selection of the applied dictionary in each iteration, we generally use \emph{n-grams} and order them according to the number of appearances in the input data. However, in future work we aim to improve this way of proceeding and apply a more sophisticated feature selection. Concerning this, quite some research has already been done. First of all, there are various metrics out there for a selection of an appropriate number of features. Regarding text classification, a valuable overview is for instance given by \cite{feature_selection}. Moreover, \cite{preprocessing} investigate the influence of different types of preprocessing applied on textual input data. 

Furthermore, \cite{optimal_vocab} explore the selection of the vocabulary in more detail and aim to find an optimal subset by providing a variational vocabulary dropout. However, this approach is computationally quite demanding and probably not suited for very large data sets.
Similarly, \cite{game_theoretic_vocab_selection} incorporate ideas from cooperative game theory with the aim to find an optimal subset of the vocabulary maximizing the performance of a classification model.\\

Another crucial point we want to address in more detail in future work is the \emph{class imbalance problem} that has an important influence in particular in the context of our use case from insurance business. Up to now, it has been a satisfying solution for our collaboration partner to summarize the smaller classes into a few super-classes and differentiate between them. However, it would also be interesting to make a more granular distinction and also in the currently applied setting with only a few considered classes we have to deal with imbalanced data to some extent. An extensive study on this topic has been done for instance by \cite{imbalance_problem} as well as \cite{imbalanced_data} and common methods to handle imbalanced data are summarized for instance by \cite{handling_imbalanced_data}.

On the other hand, \cite{hierarchical_text_classification} introduce a text classification approach that does not require any labelled data. Instead of human-labelled documents, they rather consider the description and more importantly the relationships with other categories for classification which makes this approach especially suited for data sets with a lot of different (small) classes as present in our use case. So, incorporating this ideas might also be an interesting direction for future work.\\

Finally, regarding general trends in rule learning, \emph{RIDDLE} by \cite{riddle} has to be mentioned. They bridge deep learning and rule induction resulting in a \emph{white-box} method that apparently yields state-of-the-art results in many classification tasks in rule induction. Although they claim that "the trained  weights  have  a  clear  meaning  concerning the decisions that the model takes", the level of explainability is probably still lower than the one achieved by the classical rule induction methods like \ripper\ for instance.  Moreover, for comparison we applied our approach also on the \emph{Breast Cancer} data set from the \emph{UCI machine learning repository} which has been used by \cite{riddle} in the empirical evaluation and achieved an accuracy of $95,99$\% with \foil\ and $96,55$\% using \ripper\ as opposed to $94,86$\% as mean of 5 independent repetitions using the publicly available implementation of the algorithm\footnote{See \url{https://git.app.uib.no/Cosimo.Persia/riddle}}.

\section{Methodology}
\label{Methodology}

After motivating the ideas behind this paper and summarizing related work as well as previous work on which this paper is build upon, we will introduce the applied methodology in this section.
Simply put, our iterative approach is based on a chosen rule learning method and aims to refine the generated rules according to a chosen \emph{Value of Confidence} that we define as follows.

\subsection{Value of Confidence}

\begin{definition}
\label{voc}
The \textbf{Value of Confidence} is a measure of reliability of a rule generated by a rule learning method. This numeric value is calculated on a validation data set distinct from the training set that is used to generate the rule. There are various possible calculation methods depending on the exact goal of the use case of interest. However, a common metric applied in this context might be the \emph{precision} that is also used within our experiments since it is especially important for our use case from insurance business. So, for instance one option to compute the Value of Confidence is as follows.
\begin{align*}
VoC = \frac{p}{p+n},
\end{align*} 
where $p$ is the number of positive examples and $n$ the number of negative examples covered by the rule.
\end{definition}

Note that we prefer to obtain no prediction at all rather than a wrong prediction in our use case because every bill that can be processed automatically is a gain for the company as long as we can guarantee with a very high percentage that the predicted class is correct. As a result, the precision is an appropriate Value of Confidence for our purpose. In different scenarios it might be advantageous to obtain a (possibly) wrong prediction over returning no prediction at all. For instance, if the processing of an example by a human or a different kind of method is very cost-intensive (compared to the expenses resulting from a wrong prediction), it might be bearable to obtain a wrong classification now and then. Furthermore, it might be possible that the outcomes of the rule learners are only used as decision guidance for a human. Especially in such a scenario it would be unfavourable to obtain no predictions.
A more detailed investigation of different metrics in this context will be done in future work.

\subsection{Iterative Approach}

The basic procedure of the iterative approach introduced in this paper is illustrated by the pseudo-code in Algorithm~\ref{alg:algorithm} and explained in the following.

\begin{algorithm}[b!]
    \caption{Pseudo-Code for Iterative Approach}
    \label{alg:algorithm}
    \textbf{Input}: Training and validation set\\
    \textbf{Parameter}: Maximal number of iterations, Threshold, Initial size of dictionary\\
    \textbf{Output}: Rule with corresponding Value of Confidence
    \begin{algorithmic} 
    	\STATE Restrict training data to chosen dictionary\_size
    	\STATE {$iteration \gets 0$}
        \WHILE {$iteration < \text{max\_iterations}$}
        \STATE {$rule \gets$ apply chosen rule learning method}
        \IF {$VoC(rule) < $ threshold}
        	\STATE {add false positives from validation set to training set}
        	\STATE {$\text{dictionary\_size } *= 2$}
        	\STATE {adapt data to new dictionary\_size}
        	\STATE {$iteration~+=1$}
        \ELSE
        	\STATE {return rule with corresponding VoC}
        \ENDIF
        \ENDWHILE
    \end{algorithmic}
\end{algorithm}

In a first step the given data set is split into a train, a test and a validation data set. For instance, in our experiments we use a 80/20 train-test-split and use 15\% of the training data for validation.\\
The train and validation data serves as input for our approach. As already mentioned above, the train data is used to learn a rule while the corresponding Value of Confidence is afterwards computed on the validation data.

However, before learning the first rule, the size of the input data is restricted to the chosen \emph{initial dictionary size}. Note that in our experiments we applied the \emph{TfidfVectorizer}\footnote{See \url{https://scikit-learn.org/stable/modules/generated/sklearn.feature_extraction.text.TfidfVectorizer.html}.} with a n-gram range of one to three on the raw text data for preprocessing where we considered all words that appear at least 5 times in the data set. The resulting total number of features is our original dictionary size and we have ordered the features according to the \emph{inverse document frequency}. It has shown that a reasonable value for the initial dictionary size applied in our algorithm is an eighth of the original dictionary size. This choice is small enough to significantly decrease the necessary memory consumption for the rule generation while it still covers the most important words and groups of words. Moreover, we do not want to apply a huge number of iterations but rather stop after about 5 iterations as done in our experiments since each iteration involves learning a rule which can be quite time-consuming. Using the suggested initial dictionary size, we consider the whole feature set in the fourth iteration and stop after one more iteration. It is probably not possible to find a general optimal value here, since it strongly depends on the underlying data. For instance, considering a data set where very few key words are sufficient to differentiate a large part of the data, the initial dictionary size can be chosen very small whereas a data set consisting of very similar classes might benefit from a larger initial size.

After preparing the data, we proceed as follows until the maximal number of iterations is reached or a rule of satisfying quality (with respect to the VoC) is found.
\begin{enumerate}
\item The chosen rule learning method is applied on the current training data in order to learn one rule. 
\item The Value of Confidence is computed for this rule considering the validation data.
\item The \emph{quality/reliability} of the rule is checked:
\begin{enumerate}
\item If the corresponding Value of Confidence is higher than the threshold that we pass as a parameter to the algorithm, we store the rule and remove the covered positive examples from the training set as usually done in rule learning.
\item Otherwise, we increase the dictionary size (usually we multiply it by 2) and add the covered negative examples (i.e., the false positives) from the validation set to the training set.
\end{enumerate}
\item If the quality of the rule is not satisfying, we start the next iteration considering the new training data with increased dictionary size.
\end{enumerate}

The procedure explained above and outlined in Algorithm~\ref{alg:algorithm} eventually yields one rule together with the corresponding Value of Confidence. It is repeated until a given number of rules has been generated. Additionally, we include early stopping meaning that no more rules are generated if the quality (i.e., Value of Confidence) of $n$ consecutive rules is not satisfying, where $n$ as well as the threshold determining the desired level of quality can be chosen via parameters.

In the first place, our iterative approach is intended to make it possible for common rule learning methods to better handle large/complex text-based data sets and reduce memory consumption. However, the basic idea (without increasing the feature space in each iteration) is also suitable for any other kind of data and yields improved results as shown in Section~\ref{Evaluation}.

\section{Experimental Evaluation}
\label{Evaluation}

In this section, we evaluate the iterative approach introduced in this paper on several common benchmark data sets not only from the field of text classification but also on non-textual data showing its versatile applicability. Moreover, we investigate a practical example from insurance industries.

\subsection{Experimental Setup}

As a first step, the data sets explained in the following are split into train, validation and test data. When not stated differently, we use 80\% of the input data as training data and the remaining 20\% for testing. From the training data we use 15\% as validation data set for the application of our iterative approach. This additional split is not necessary when we use the ordinary method. So, the corresponding outcomes presented in the comparison in Section~\ref{Results} are obtained by considering the whole training data (i.e., 80\% of the total input data) without generating a separate validation set.
Note that at this point preprocessing has already been done. So, in particular for the considered text-based data sets, the textual information has already been transformed into binary vectors where the attributes are ordered according to the \emph{inverse document frequency} as already mentioned above.

Before starting with our approach, we define a \emph{start dictionary size} which is usually an eighth of the total number of attributes as explained above. Regarding the maximal number of iterations and the applied threshold for the \emph{Value of Confidence}, we always apply the same settings; namely at most 5 iterations with a threshold of $0.9$. However, note again that we add the rule resulting from the last iteration to our set of rules independent of the corresponding \emph{Value of Confidence}. So, in the final ruleset there might be rules with an unsatisfying reliability but we can ignore them during evaluation. In fact, we are interested in the differences that can be observed by applying only rules with a certain reliability as further shown in Section~\ref{DetailedAnalysis}.

After that, we can define the rule learning method we want to apply as well as the number of rules that should be generated and our iterative approach proceeds as explained in Section~\ref{Methodology}.

Before going into detail on the obtained results, we briefly explain the underlying data considered in our experiments. We start with the considered benchmark data sets and discuss the results obtained on them in Section~\ref{Results}. Afterwards, in Section~\ref{UseCase}, we will focus on our use case from insurance industries showing that the benefits achieved by our iterative approach are not only present considering some standard benchmark data sets but also on a use case of crucial importance to our industrial collaboration partner.

\subsubsection{Hatespeech}

This data set from Kaggle\footnote{See \url{https://www.kaggle.com/datasets/mrmorj/hate-speech-and-offensive-language-dataset}.} consists of about 25 thousand Twitter posts labelled as \emph{hate speech}, \emph{offensive language} or \emph{neither}. In our experiments we summarized the first two classes into one in order to differentiate simply between \emph{Hate Speech/ Offensive Language} or not. So, in our case this is a binary classification task.
After preprocessing we consider about 8000 attributes representing the occurrence of words/word groups like \emph{hate, dumb, monkey} as well as a lot of swearwords we do not want to mention here. A simple rule learned in this context could be for instance

\begin{lstlisting}[style=Prolog]
		IF	dumb = 1
		THEN	Type = Hate Speech

\end{lstlisting}

meaning that a tweet should be considered as \emph{Hate Speech} if the word \emph{dumb} appears. Of course, there are also more complex rules not just considering the presence of one certain swear word because some words can be used in a completely different context. For example, the word \emph{monkey} is sometimes used in a racist context but also in innocent tweets about a zoo visit resulting in rules like

\begin{lstlisting}[style=Prolog]
		IF	monkey = 1
		AND cute = 1
		THEN	Type = NOT Hate Speech.

\end{lstlisting}

\subsubsection{Reuters}

There are various variants of this data set commonly used in literature. We considered the version contained in the python \emph{nltk} package\footnote{See \url{https://www.kaggle.com/datasets/boldy717/reutersnltk}.} consisting of 10788 news documents assigned to the according categories. After preprocessing, the data set comprised nearly 11 thousand attributes eventually resulting in rules like the following.

\begin{lstlisting}[style=Prolog]
		IF	water = 1
		AND carry = 1
		THEN	Type = SHIP

\end{lstlisting}

Note that we distinguish between the 10 most common categories while summarizing the remaining smaller classes as \emph{other}.

\subsubsection{IMDB}

This data set from Kaggle\footnote{See \url{https://www.kaggle.com/datasets/lakshmi25npathi/imdb-dataset-of-50k-movie-reviews}.} contains 50 thousand informal movie reviews from the \emph{ Internet Movie Database} mostly used for sentiment analysis. After preprocessing, we have more than 70 thousand attributes available. It has shown that \foil\ is able to handle this amount of features while \ripper\ is not able to do so due to its increased complexity resulting in extensive memory consumption. So, for our experiments with \ripper\ we cropped the feature space and considered only the 20 thousand most important words according to the \emph{inverse document frequency}.
An example of a learned rule in this context is as follows.

\begin{lstlisting}[style=Prolog]
		IF	bad = 1
		AND great = 0
		AND like = 0
		THEN	Type = negative

\end{lstlisting}

\subsubsection{Non-textual data sets}

Beside these text-based data sets, we also considered non-textual input data in order to investigate the advantages achieved just by assigning a \emph{Value of Confidence} to each generated rule aiming to maximize this value in our iterative approach without the need of restricting the data to a certain dictionary size. More precisely, we considered the following data sets discussed in more detail in the Supplementary Material of our previous work.\footnote{See \url{https://arxiv.org/pdf/2311.07323}.}

\begin{enumerate}[(i)]
\item Spambase\footnote{See \url{https://archive.ics.uci.edu/ml/datasets/spambase}.}
\item Heart Disease\footnote{See \url{https://archive.ics.uci.edu/dataset/45/heart+disease}.}
\item Car Evaluation\footnote{See \url{https://archive.ics.uci.edu/dataset/19/car+evaluation}.}
\item Diabetes\footnote{See \url{https://www.kaggle.com/datasets/uciml/pima-indians-diabetes-database}.}
\item Breast Cancer\footnote{See \url{https://archive.ics.uci.edu/dataset/15/breast+cancer+wisconsin+original}.}
\end{enumerate}

\subsection{Objectives \& Summary}
\label{Results}

The empirical evaluation of the iterative approach introduced in this paper in particular sought to answer the following questions.

\newcommand*{\RQ}[1]{\textbf{RQ{#1}}}
\makeatletter 
\let\orgdescriptionlabel\descriptionlabel
\renewcommand*{\descriptionlabel}[1]{%
  \let\orglabel\label
  \let\label\@gobble
  \phantomsection
  \protected@edef\@currentlabel{#1\unskip}%
  \let\label\orglabel
  \orgdescriptionlabel{#1}%
}
\makeatother

\begin{description}
\item[\RQ 1 \label{q1}]\emph{Accuracy compared to the base method.}
  Can the iterative approach provide better accuracy 
  of classification prediction than the base method, i.e. the ordinary rule learning method.

\item[\RQ 2 \label{q2}]\emph{Memory consumption compared to the base method.}
  Is our iterative approach able to significantly reduce the memory consumption for rule generation compared to the ordinary method.
  
\item[\RQ 3 \label{q3}] \emph{Industrial case study.}
  Are the advantages regarding classification accuracy and memory consumption also observable for the classification of
  dental bills, an industrial use case.
  
\item[\RQ 4 \label{q4}] \emph{Level of reliability.}
  What is the impact of the \emph{Value of Confidence} as a metric of reliability concerning classification accuracy? 
\end{description}

In order to investigate these questions, we consider the above-mentioned data sets.
Note that the reported results are always obtained on the test data.

\begin{table}[b!]
\centering
\begin{tabular}{l|l|r|r}
\hline
                                                                                                                                       \hfil \textbf{Data} \hfill & \hfil \textbf{Learner} \hfill & \hfil \textbf{Memory Consumption in GiB} \hfill & \textbf{Accuracy in \%} 
  \\ \hline\hline
Hatespeech & \foil\ & $6,72$  & $82,00$ \\
 & \foil\ - iter. & $3,92$ & $86,44$\\
 & \ripper\ & $30,54$ & $89,30$ \\
 & \ripper\ - iter. & $13,45$ & $92,64$ \\
\hline
Reuters & \foil\ & $4,27$ & $72,21$ \\
 & \foil\ - iter. & $2,92$  & $81,74$  \\
 & \ripper\ & $13,19$  & $78,89$ \\
 & \ripper\ - iter. & $11,26$  & $80,49$\\
\hline
IMDB & \foil\ & $148,80$ & $79,13$ \\
& \foil\ - iter. & $107,57$ & $79,31$ \\
& \ripper\ & $113,60$\footnote{Note that a smaller feature space has been used for the application of \ripper.}  & $68,39$ \\
& \ripper\ - iter. & $91,92$  & $75,01$\\
\hline
\end{tabular}
\vspace{2ex}
\caption{Performance of our approach on different benchmark problems for text classification. Note that \emph{iter.} denotes the iterative approach introduced in this paper.} \label{tab:benchmark_results_text}
\end{table}

\begin{figure}
\caption{Illustration of Accuracies shown in Table~\ref{tab:benchmark_results_text}.}
\label{fig:acc_textual}  
\begin{center}
  \setlength{\fboxsep}{0pt}
  \fbox{
    \includegraphics[width = \textwidth]{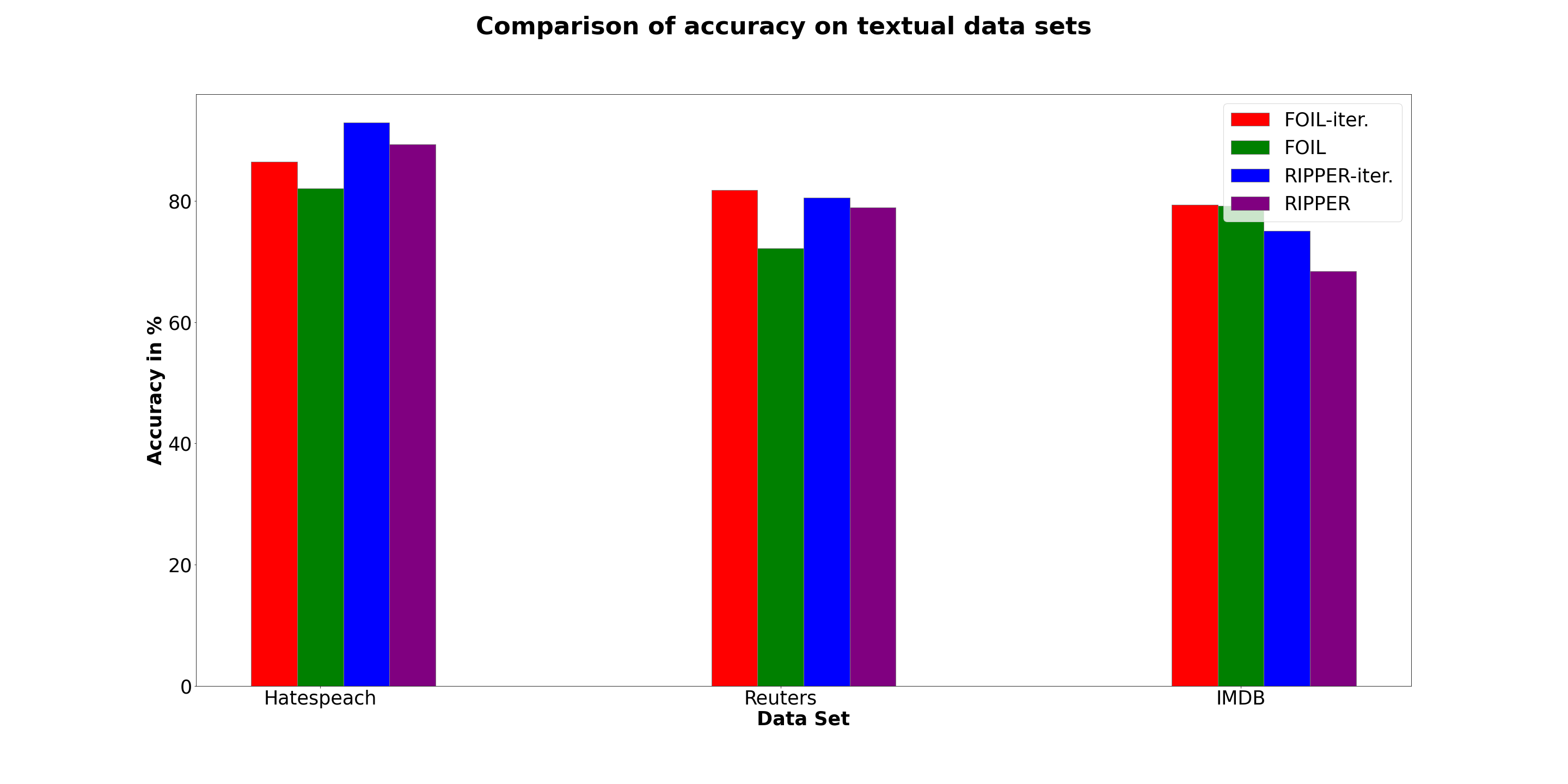}
  }
\end{center}
\end{figure}

\begin{figure}[h!]
\caption{Illustration of Memory Consumptions shown in Table~\ref{tab:benchmark_results_text}. Note that the memory consumption illustrated for \ripper\ applied on the \emph{IMDB} data set corresponds to a reduced feature space compared to the application of \foil.}
\label{fig:mem_textual}  
\begin{center}
  \setlength{\fboxsep}{0pt}
  \fbox{
    \includegraphics[width = \textwidth]{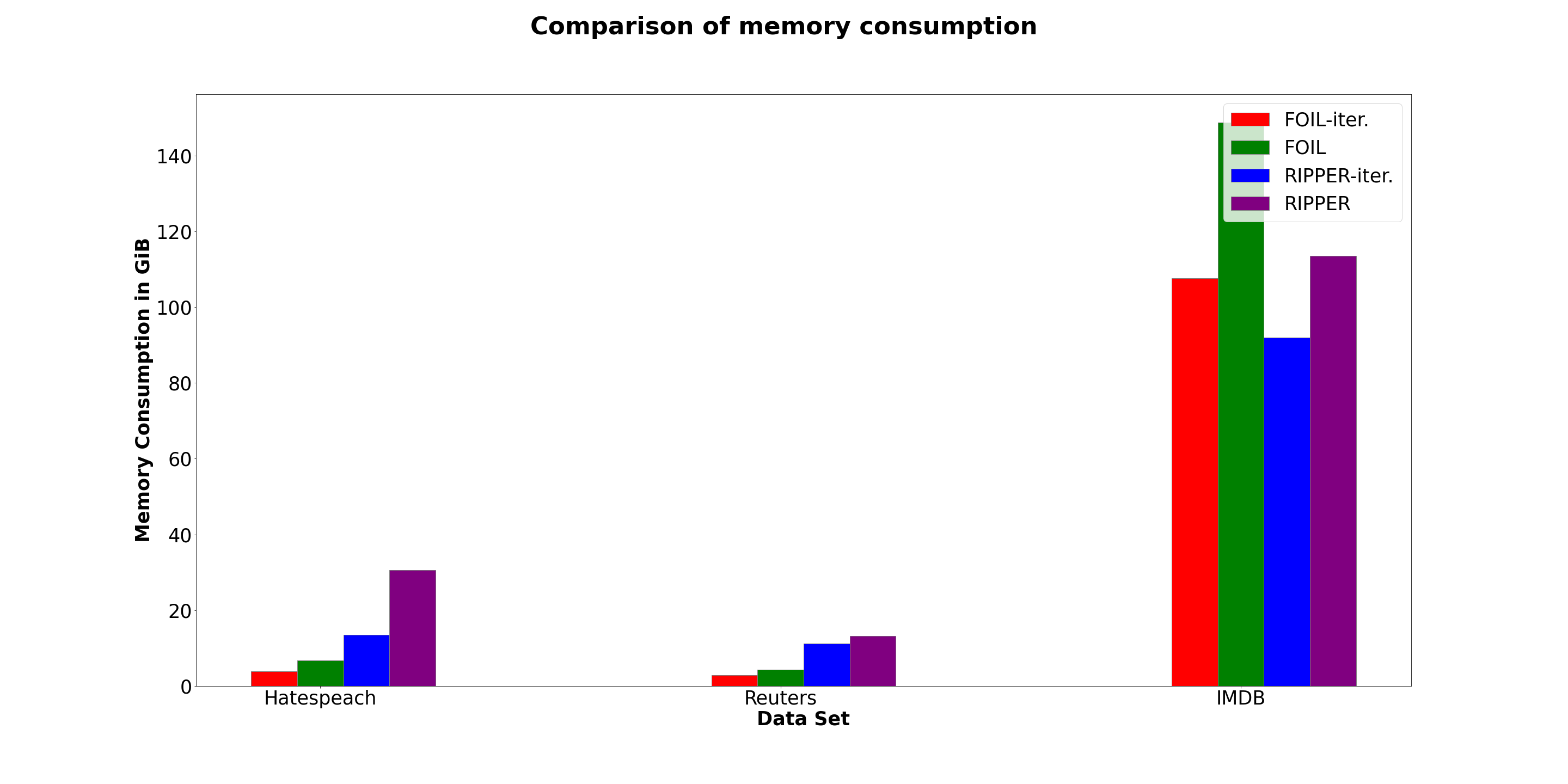}
  }
\end{center}
\end{figure}

Concerning the textual data sets we not only compare the resulting accuracy from our proposed iterative approach with the ordinary method but also the memory consumption measured in our experiments. The corresponding results are shown in Table~\ref{tab:benchmark_results_text} and visualized in Figure~\ref{fig:acc_textual}~and~\ref{fig:mem_textual}, respectively. Note that all of the experiments are performed on a \emph{AMD Ryzen Threadripper 2950X WOF} CPU.

With respect to accuracy, we can clearly observe that our iterative approach outperforms the ordinary method on the considered data sets for both \foil\ and \ripper. The only exception is the application of \foil\ on the \emph{IMDB} data set, where both approaches are equivalent. A possible reason for that might be the kind of language used in this data set which could also explain the generally rather poor performance of \ripper\ on this example (beside the already mentioned restriction of the feature space). The \emph{IMDB} data set consists of movie reviews written in simple language often using abbreviations and containing typographical errors. This might have a significant influence on the dictionary we use for rule learning. In future work we aim to improve the preprocessing of the text-based input data by applying large language models, for instance. Regarding this, \cite{proofread} have recently introduced a very promising approach to fix errors in a text document.

Furthermore, concerning memory consumption it is clearly visible that we are able to significantly reduce the memory consumption by applying the way of proceeding introduced in this paper. Especially using the \foil\ algorithm, we can observe that the memory consumption is reduced by about a third on all of the considered benchmarks. Using \ripper, it seems that the reduction of memory consumption rather depends on the underlying data. While we notice a remarkable reduction of more than a half on the \emph{Hatespeech} data set (where the two classes are mostly distinguishable by considering the occurrence of some swear words), the reduction of the memory consumption on the other two benchmark data sets is not that distinct but still clearly visible with about $20$\%.

\begin{table}[t]
\centering
\begin{tabular}{l|r|r|r|r|r}
\hline
                                                                                                                                        & \hfil \textbf{Spambase} \hfill & \textbf{Heart Disease} & \hfil \textbf{Car} \hfill & \hfil \textbf{Diabetes} \hfill  & \hfil \textbf{Breast Cancer} \hfill
  \\ \hline\hline
\foil\ & $87,69$  & $81,95$  & $92,00$ & $92,94$ & $96,00$\\
\hline
\foil\ - iter. & $89,71$  &  $85,29$ & $95,07$ & $94,80$  & $95,99$\\
\hline
\ripper\ & $92,18$ & $82,16$ & $92,47$ & $88,84$ & $95,08$ \\
\hline
\ripper\ - iter. & $91,74$ & $82,78$ & $93,84$  & $ 90,49$ &  $96,55$ \\
\hline
\end{tabular}
\vspace{2ex}
\caption{Accuracy in \% achieved by our approach on different non-textual benchmark problems. Note that \emph{iter.} denotes the iterative approach introduced in this paper.} \label{tab:benchmark_results_nominal}
\end{table}

\begin{figure}[b!]
\caption{Illustration of Accuracies shown in Table~\ref{tab:benchmark_results_nominal}.}
\label{fig:acc_nominal}  
\begin{center}
  \setlength{\fboxsep}{0pt}
  \fbox{
    \includegraphics[width = \textwidth]{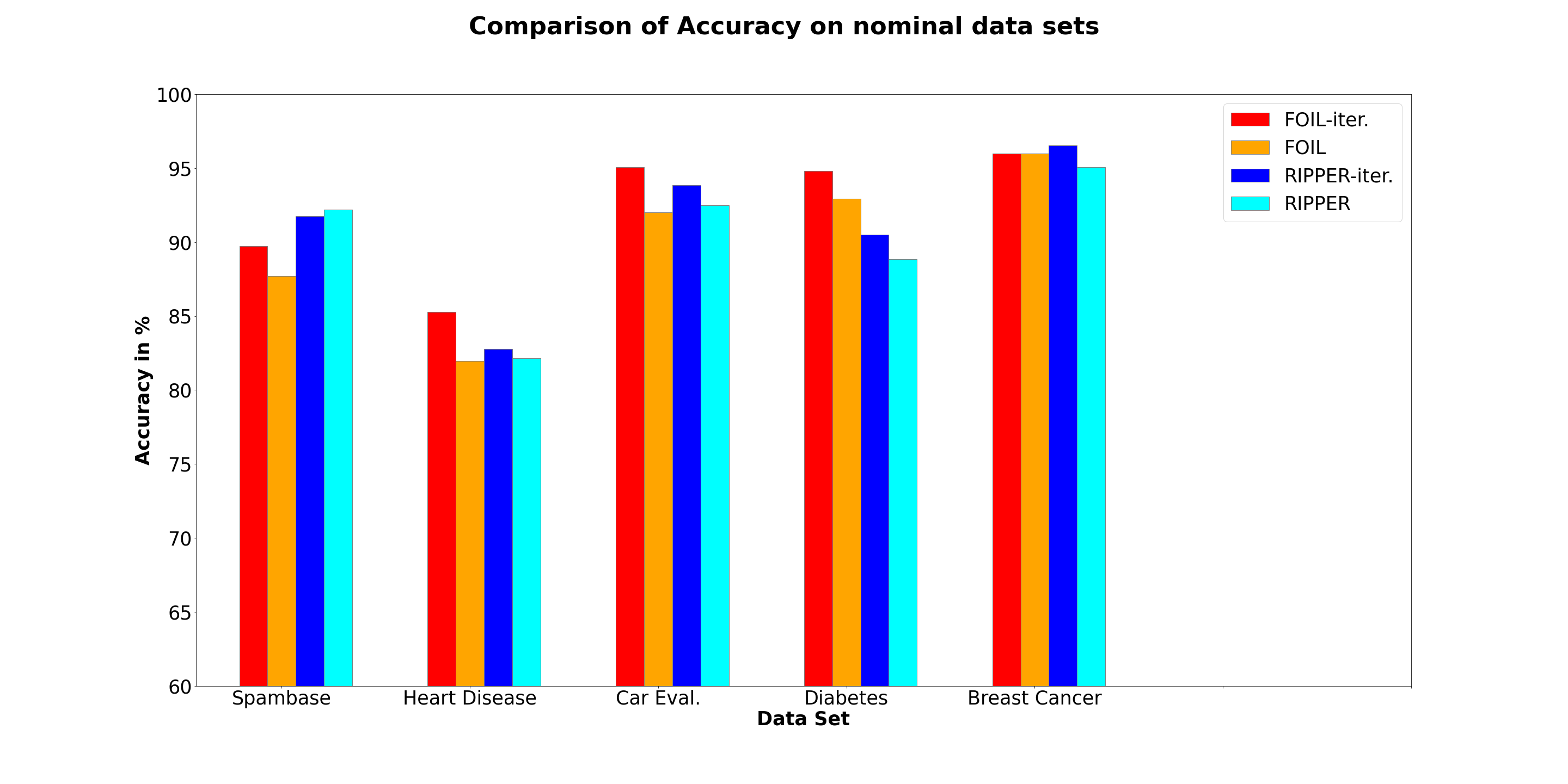}
  }
\end{center}
\end{figure}

Regarding time consumption, we did not investigate the differences between the two approaches in that detail but in general we observed an increased time consumption when \ripper\ is applied within our approach, while our iterative approach could even reduce the run time using \foil. For instance, on the \emph{Hatespeech} data set using \foil\ we observed a total time consumption of about $37$ minutes compared to approximately $77$ minutes corresponding to the classical method. On the other hand, applying \ripper\ results in a total time consumption of about $19$ hours compared to about $4$ hours with the classical method. However, note that at the end of the day the introduced iterative approach is intended to extend our framework for a versatile application of rule learning methods we already established in previous work. In particular, in combination with the modular approach proposed in ~\cite{modular_approach} the total time consumption can be reduced by a multiple when we apply parallelization. In order to do so, the reduced memory consumption achieved by the iterative approach introduced in this paper is extremely beneficial.

In addition, we also evaluate our iterative approach on some nominal data sets as mentioned above. The corresponding accuracy is depicted in Table~\ref{tab:benchmark_results_nominal} and Figure~\ref{fig:acc_nominal}. As clearly visible, our approach yields also significantly improved results on most of the considered non-textual benchmarks and outperforms the classical method by up to $3,3$\%. 

So, all in all, we can positively answer Questions~\ref{q1}~and~\ref{q2}.

\subsection{Use Case: Reimbursement}
\label{UseCase}

The \emph{Allianz Private Krankenversicherung (APKV)} is an insurance company offering health insurance services in Germany. As already mentioned, the inspiration for this work stems from a use case we worked on during a collaboration with this company.
In our previous work (\cite{modular_approach, voting_approach}), we have already described the use case at hand in detail. However, summed up, an insurance company regularly receives bills handed in by the clients asking for reimbursement. Automated processing of these bills is desired in order to lower costs and to gain an edge over the competition by reducing the time until the client receives the reimbursed money.

As decision making, in particular in this sensitive area, should be transparent to both parties, the operational use of black-box machine learning algorithms is often seen critically by the stakeholders and is in many cases avoided. As a consequence, rule learning achieving a comparable performance offers the desired advantage of explainability.  

For our case study, we are focusing on \emph{dental bills}. On those bills, the specific type of dental service per row on the bill is unknown but needed for deciding on the amount of refund. Especially differentiating between material costs and other costs is of crucial importance. 

\begin{table}[t]
\centering
\begin{tabular}{l|r|r|r|r|r}
\hline
                                                                                                                                       \hfil \textbf{Learner} \hfill & \hfil \textbf{Memory (GiB)} \hfill & \hfil \textbf{Threshold} \hfill & \hfil \textbf{Predicted} \hfill & \hfil \textbf{Correct} \hfill & \hfil \textbf{Precision (\%)} \hfill
  \\ \hline\hline
 \foil\ &  $204,53$  &  & $178.910$ & $137.564$  & $76,89$ \\
 \foil\ - iter. & $126,78$ & $0$ & $232.476$ & $175.515$ & $75,50$ \\
  &  & $0,6$ & $155.634$ & $142.798$ & $91,75$ \\
 &  & $0,7$ & $150.601$ & $140.339$ & $93,19$\\
 &  & $0,8$ & $144.099$ & $135.844$ & $94,27$ \\
 &  & $0,9$ & $119.635$ & $115.697$ & $96,71$ \\
 \hline
  \ripper\ & $213,82$  &  &  $106.230$ & $92.041$ & $86,64$\\
  \ripper\ - iter. & $165,37$ & $0$ & $150.538$ & $135.590$ & $90,07$  \\
  &  & $0,6$ & $150.538$ & $135.590$ & $90,07$ \\
 &  & $0,7$ & $149.166$ & $134.722$ & $90,32$ \\
 &  & $0,8$ & $141.878$ & $129.067$ & $90,97$ \\
 &  & $0,9$ & $84.815$ & $79.702$ & $93,97$ \\
\hline
\end{tabular}
\vspace{2ex}
\caption{Performance of our approach on the reimbursement case study concerning dental bills. Note that \emph{iter.} denotes the iterative approach introduced in this paper and the \emph{Threshold} corresponds to the \emph{Value of Confidence} of each rule meaning that rules with a reliability below the threshold are ignored.} \label{tab:results_allianz}
\end{table}

\begin{figure}[b!]
\caption{Illustration of Memory Consumption \& Accuracy shown in Table~\ref{tab:results_allianz}.}
\label{fig:results_allianz}  
\begin{center}
  \setlength{\fboxsep}{0pt}
  \fbox{
    \includegraphics[width = \textwidth]{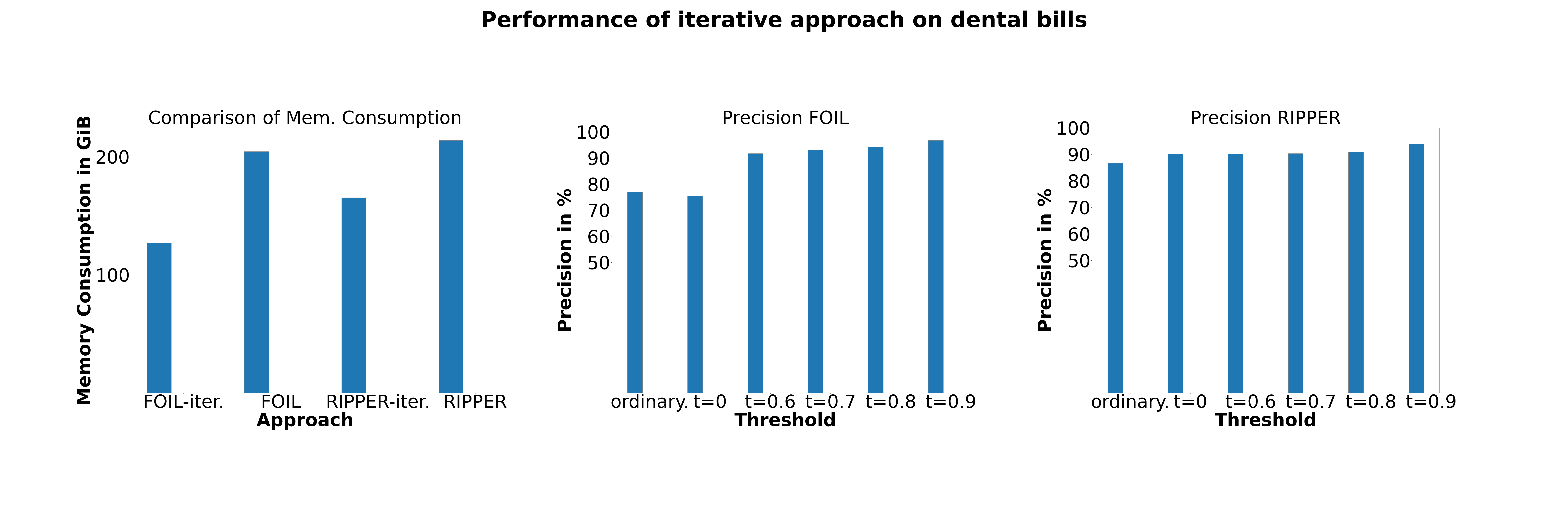}
  }
\end{center}
\end{figure}

In collaboration with the APKV, we have been provided with an anonymized training data set consisting of nearly one million instances.
As opposed to our previous work, where we only considered structured information on the bills such as cost, date and simple engineered features, in this paper we especially aim to work with the textual data and make predictions based on the occurrence of certain words or word groups where we had to restrict to the 8000 most common ones using \foil\ and the 3000 most common ones for \ripper\ due to the extensive memory consumption. 

Originally, large language and transformer models such as \emph{RoBERTa} (\cite{Liu2019RoBERTaAR}) have been applied to process the bills. Due to pending non-disclosure agreements we cannot go into detail about the exact procedure\footnote{For more information please directly contact \href{mailto:gabriela.dick_guimaraes@allianz.de}{gabriela.dick\_guimaraes@allianz.de}.} but at the end of the day these highly complex methods have been applied on a combination of both the textual information as well as the engineered features mentioned above. Considering exclusively the textual information has not been tested yet.

However, in order to investigate the benefit of applying our approach on real-world text data, we considered the textual information exclusively in our experiments. Taking also engineered features into account is left to future work, where we want to bring everything together and apply a combination of all three of our introduced approaches (modular, voting and iterative) on all available features.

In the experiments conducted during the evaluation of our approach on the industrial use case, we especially considered the precision of the fully satisfied rules and did not apply partial matching (cf.~\cite{partial_matching}) as usually done during evaluation. So, in case no rule is completely satisfied for a considered example we do not make a prediction instead of additionally checking how many of the conditions of each rule are fulfilled and predict the label corresponding to the rule with the highest percentage of satisfied conditions. 

Summed up, by considering the results shown in Table~\ref{tab:results_allianz} and Figure~\ref{fig:results_allianz} we can answer Question~\ref{q3} as follows. Both the reduction of the memory consumption as well as the increase of classification accuracy are also clearly visible on the industrial use case on dental bills. More precisely, considering \foil\ we can almost half the memory consumption and concerning the precision of the applied rules, the positive effect of the introduced \emph{Value of Confidence} is clearly visible. While the precision of our iterative approach without restrictions to the reliability of the applied rules is slightly smaller than the one achieved by the classical method, the application of a threshold in this context immediately improves the results enormously. Using a threshold of $0.6$ already yields a precision (i.e., number of correctly predicted examples divided by the total number of examples where a prediction has been made) of nearly $92$\% correctly predicting even more examples than the classical method. Further restricting the reliability of the applied rules and using a threshold of $0.9$ yields a precision of almost $97$\%, while still predicting correctly about 115 thousand examples which corresponds to about half of the test examples. At the end of the day, this means that our approach makes it possible to classify half of the dental bills in an automated manner with an extremely high accuracy and -- what is even more important -- the resulting predictions are fully explainable.

Considering \ripper\ we observe very similar results reducing the memory consumption by about a third and increasing the precision from $86.64$\% achieved by the classical method to up to $94$\% obtained by our iterative approach using a threshold of $0.9$. Note that the corresponding experiments have been conducted with the general restriction to learn at most 10 rules for each label for both approaches. However, the ordinary method returned only 2-3 rules for 8 of the 10 labels due to the integrated early stopping according to the \emph{description size} -- a measure of total complexity of the model aiming to balance between minimization of classification error and minimization of model complexity. Using the same amount of rules with our iterative approach, we can correctly classify $75401$ examples from $83222$ examples where one rule is satisfied. This corresponds to a precision of $90.60$\% independent of the chosen threshold meaning that the generated rules all have a Value of Confidence of more than $0.9$. Nevertheless, we decided to present the results of the 10 rules learned for each label using our iterative approach in Table~\ref{tab:results_allianz} and Figure~\ref{fig:results_allianz} because on the one hand this shows that the applied early stopping in the classical approach can sometimes be too restrictive and, on the other hand, it allows a deeper insight in the effect of applying a threshold concerning the Value of Confidence on the rules used for evaluation.

So, all in all, our iterative approach outperforms the classical approach also on the industrial use case concerning both classification accuracy as well as memory consumption.

Moreover, note that the derived rules are of great use, even for non-automated classification of such medical bills to achieve more consistency and transparency in the decision making and gain deeper insights in the data, in general.

\subsection{Detailed Analysis}
\label{DetailedAnalysis}

\begin{table}[b!]
\centering
\begin{tabular}{l|l|l|r|r|r|r|r}
\hline
                                                                                                                                       \hfil \textbf{Data} \hfill & \hfil \textbf{Learner} \hfill & \hfil \textbf{Metric} & \hfil \textbf{$t=0$} & \hfil \textbf{$t=0.6$} \hfill & \textbf{$t=0.7$} & \textbf{$t=0.8$} & \textbf{$t=0.9$}
  \\ \hline\hline
Hatespeech & \foil\ & predicted & $4312$ & $3488$ & $3430$ & $3221$ &  $3083$ \\
 & & correct & $3641$ & $3347$ & $3303$ & $3164$  & $3047$  \\
 & & accuracy & $84,44$\% & $95,96$\% & $96,30$\% & $98,23$\% & $98,83$\% \\
 \hline
 & \ripper\ & predicted  & $4201$ & $4201$ & $4190$ & $3951$ & $3932$ \\
  & & correct & $4084$ & $4084$ & $4075$ & $3904$ &  $3886$  \\
 & & accuracy & $97,21$\% & $97,21$\%  & $97,26$\% & $98,81$\% & $98,83$\% \\
\hline
Reuters & \foil\ & predicted & $1585$ & $1498$ & $1490$ & $1477$ & $1469$ \\
&  &  correct & $1357$ & $1319$ & $1314$ & $1306$ & $1300$  \\
&  &  accuracy & $85,62$\% & $88,05$\% & $88,19$\% &  $88,42$\% & $88,50$\%\\
\hline
 & \ripper\ & predicted  & $1713$ & $1692$ & $1684$ & $1617$ & $1302$ \\
 &  &  correct & $1447$ & $1433$ & $1427$ & $1376$ & $1112$ \\
&  &  accuracy & $84,47$\% & $84,69$\% & $84,74$\% & $85,10$\% & $85,41$\% \\
\hline
IMDB & \foil\ & predicted  & $7560$ & $7545$ & $7545$ & $7185$ & $6434$ \\
&  &  correct &  $6263$ & $6252$ & $6252$ & $6009$ & $5454$ \\
&  &  accuracy & $82,84$\% & $82,86$\% &  $82,86$\% & $83,63$\% & $84,77$\% \\
\hline
& \ripper\ & predicted  & $7668$ & $7668$ & $6937$ & $3433$ & $1919$ \\
&  &  correct & $6034$ & $6034$ & $5501$ & $2846$ & $1697$ \\
&  &  accuracy & $78,69$\% &  $78,69$\% & $79,30$\% & $82,90$ & $88,43$\% \\
\hline
\end{tabular}
\vspace{2ex}
\caption{Comparison of the classification outcomes considering only rules satisfying a certain level of reliability $t$ measured by its \emph{Value of Confidence}.} \label{tab:results_thresholds}
\end{table}

As a part of this paper, we have introduced a \emph{Value of Confidence} that can be used as a metric of reliability of a generated rule. This section aims to investigate the influence of this value on the precision achieved during evaluation (cf.~\ref{q4}). 

\begin{figure}[h!]
\caption{Illustration of Accuracy regarding \foil\ shown in Table~\ref{tab:results_thresholds}.}
\label{fig:thresholds_foil}  
\begin{center}
  \setlength{\fboxsep}{0pt}
  \fbox{
    \includegraphics[width = \textwidth]{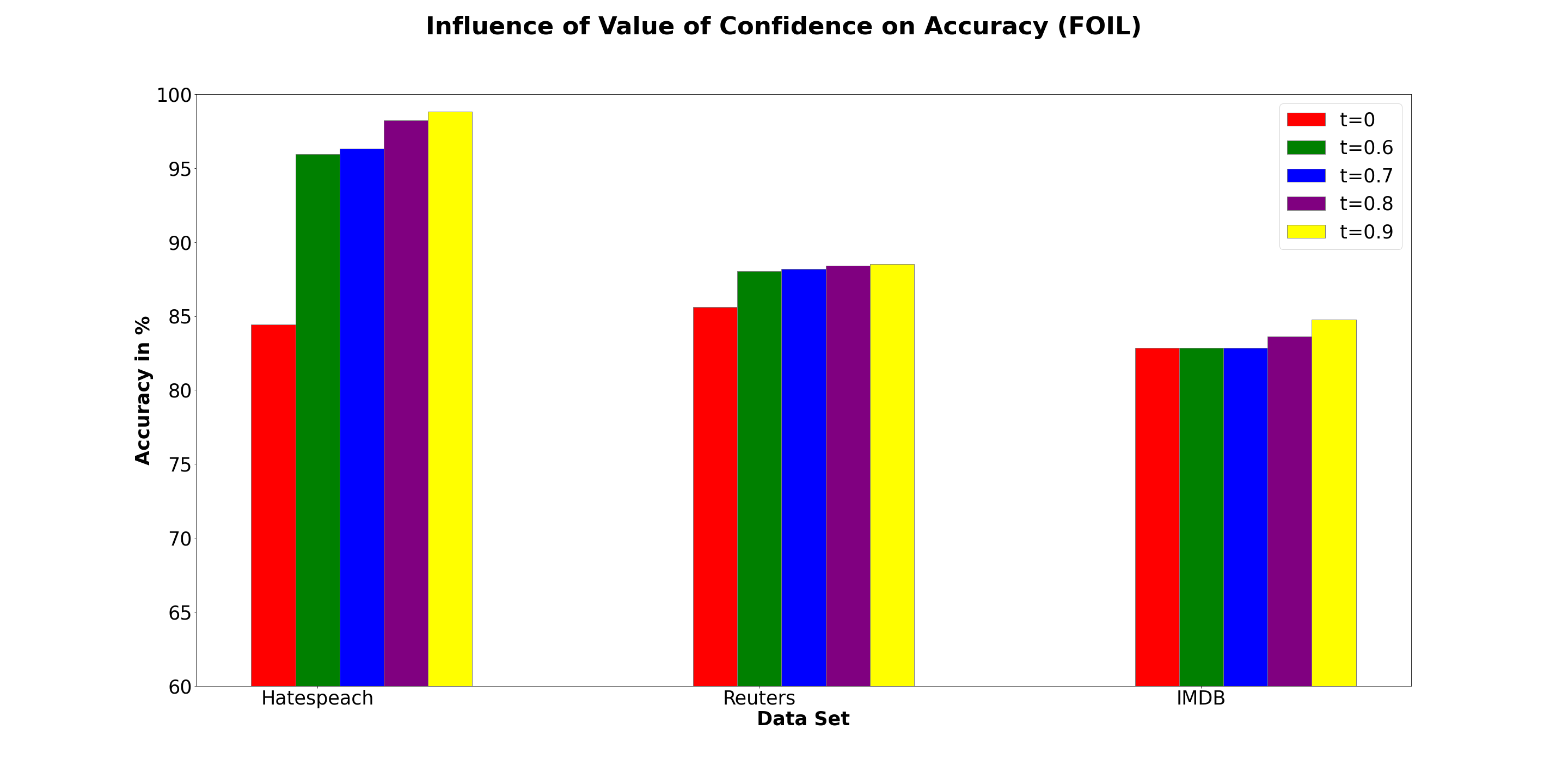}
  }
\end{center}
\end{figure}

\begin{figure}[h!]
\caption{Illustration of Accuracy regarding \ripper\ shown in Table~\ref{tab:results_thresholds}.}
\label{fig:thresholds_ripper}  
\begin{center}
  \setlength{\fboxsep}{0pt}
  \fbox{
    \includegraphics[width = \textwidth]{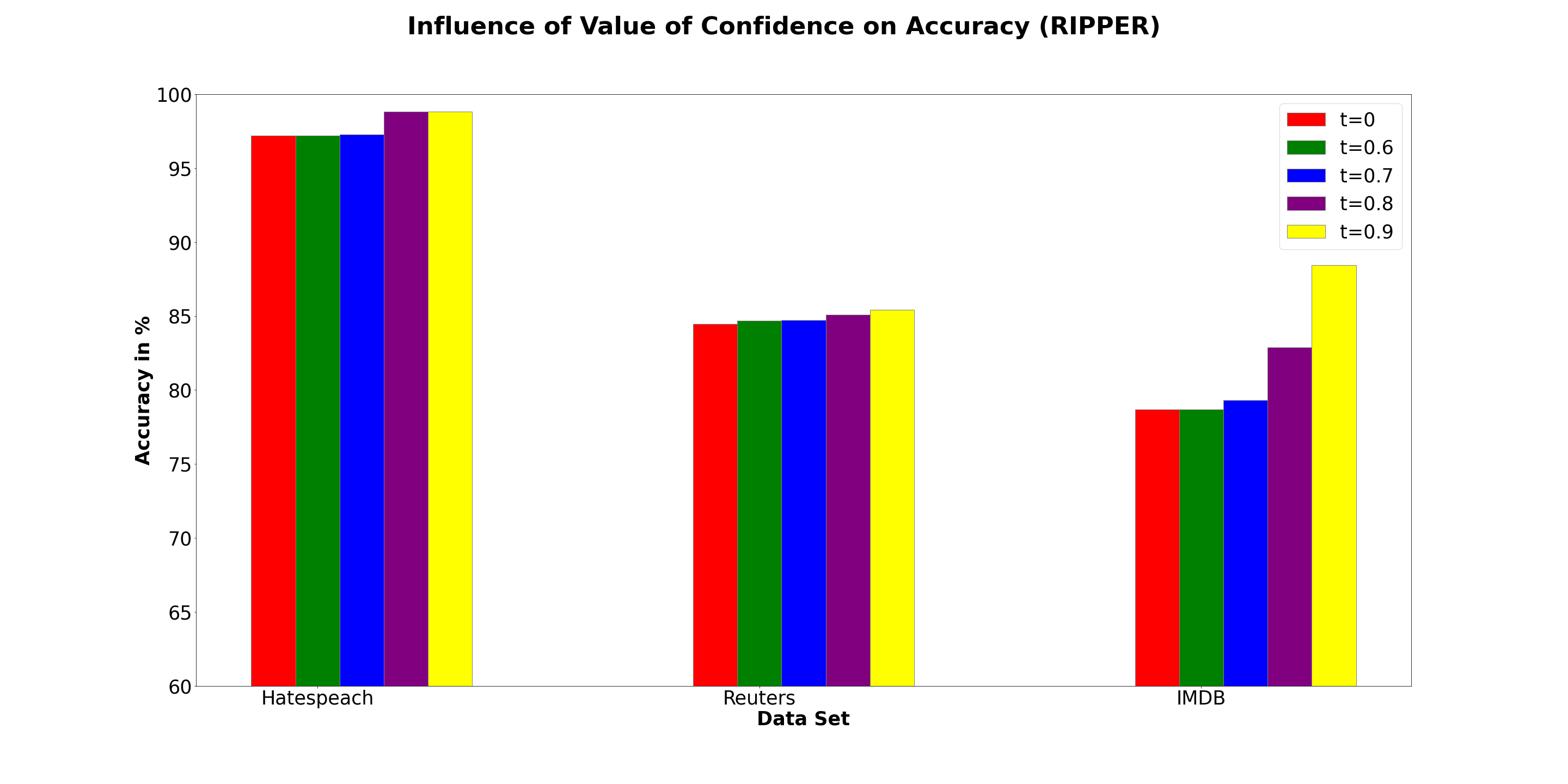}
  }
\end{center}
\end{figure}

For this purpose, we apply thresholds $t$ from 0.6 to 0.9 and consider only rules with a $\text{VoC} > t$. The corresponding results are shown in Table~\ref{tab:results_thresholds} as well as Figure~\ref{fig:thresholds_foil}~and~\ref{fig:thresholds_ripper}. In this context, we only consider fully satisfied rules and do not apply partial matching as also done and explained in Section~\ref{UseCase}. 

In order to answer question~\ref{q4}, we again illustrate for each of the considered textual benchmark data sets the number of examples where a prediction has been made (i.e., one rule is completely satisfied) together with the percentage of correctly classified examples. As expected, the number of classified examples decreases with increasing threshold and the associated reduction of total rules. However, as desired, the remaining rules are obviously more reliable and the percentage of correctly predicted examples steadily increases for both \foil\ and \ripper\ on each of the considered benchmarks. So, all in all, the incorporation of a \emph{Value of Confidence} definitely has a positive impact on the precision of the made predictions.

\section{Conclusion \& Future Work}
\label{Conclusion}

In this paper we present an extension to classical rule learning methods making use of a \emph{Value of Confidence} as metric of reliability. This novel approach is especially suited for the application of rule learners on textual input data but the iterative approach is not only beneficial for gaining more control over the applied dictionary but has shown to be also advantageous for nominal data by optimizing the reliability of the generated rules in each iteration.

By combining the approach introduced in this paper with the two approaches to rule learning we already introduced in our previous work (cf.~\cite{modular_approach,voting_approach}) we obtain a framework for explainable classifications that can be applied in various scenarios handling different types of data in a production environment.

Concerning future work, we aim to integrate a more sophisticated preprocessing applying for instance large language models to improve the choice of the dictionary. In the course of this, we will also investigate different ways of sorting the basic dictionary with the goal to find the best possible starting dictionary used in the first iteration of our approach.
Moreover, using computer vision approaches in order to incorporate the position of words in a document might be another interesting consideration we aim to investigate in future work because especially in our main use case concerning reimbursement, the considered bills are mostly standardized and the crucial information is usually located in a certain area in the document.

\bibliographystyle{plainnat}
\bibliography{literature}


\end{document}